\documentclass[letterpaper, 10 pt, conference]{ieeeconf}  

\IEEEoverridecommandlockouts                              

\usepackage{graphics} 
\usepackage{mathptmx} 
\usepackage{times}
\usepackage{epsfig}
\usepackage{graphicx}
\usepackage{amsmath}
\usepackage{amssymb}
\usepackage{bm}
\usepackage{color}
\usepackage{multirow}
\usepackage{makecell}
\usepackage{algpseudocode}
\usepackage[ruled,vlined]{algorithm2e}
\usepackage{verbatim}
\usepackage{subfig}
\usepackage{float}
\usepackage{verbatim}

\title{\LARGE PlaneSegNet: Fast and Robust Plane Estimation Using a Single-stage Instance Segmentation CNN
}

\author{
Yaxu Xie
\thanks{DFKI - German Research Center for Artificial Intelligence, Kaiserslautern, Germany. E-mail: { \{first\_name\}.\{last\_name\}@dfki.de}}

\thanks{\textbf{Acknowledgment:} This research has been partially funded by the German BMBF projects MOVEON (01IS20077) and RACKET (01IW20009).}
\and
Jason Rambach
\and
Fangwen Shu
\and
Didier Stricker
}

\begin{document}
\maketitle
\thispagestyle{empty}
\pagestyle{empty}

\begin{abstract}
Instance segmentation of planar regions in indoor scenes benefits visual SLAM and other applications such as augmented reality (AR) where scene understanding is required. Existing methods built upon two-stage frameworks show satisfactory accuracy but are limited by low frame rates. In this work, we propose a real-time deep neural architecture that estimates piece-wise planar regions from a single RGB image. Our model employs a variant of a fast single-stage CNN architecture to segment plane instances.
Considering the particularity of the target detected, we propose Fast Feature Non-maximum Suppression (FF-NMS) to reduce the suppression errors resulted from overlapping bounding boxes of planes. We also utilize a Residual Feature Augmentation module in the Feature Pyramid Network (FPN) . Our method achieves significantly higher frame-rates and comparable segmentation accuracy against two-stage methods.
We automatically label over 70,000 images as ground truth from the Stanford 2D-3D-Semantics dataset. Moreover, we incorporate our method with a state-of-the-art planar SLAM and validate its benefits.

\end{abstract}


\section{Introduction}\label{sec:intro}
Detection of 3D geometry features in scenes supports tasks such as 3D scene understanding, robot navigation and Simultaneous Localization and Mapping (SLAM). Planes, as one of the most fundamental geometry features, widely present in most man-made scenes. In indoor applications, piece-wise plane estimation benefits building modeling \cite{Wang17}, visual SLAM~\cite{Rambach19} and robot navigation~\cite{Pears01}. Planar information is also valuable for mobile Augmented Reality (AR) applications. In outdoor urban environments, plane estimation is used for object level reconstructing of buildings~\cite{Nan16} and 6-DoF pose estimation of object~\cite{Rangesh2020GroundPlane}. 

Geometry model based plane estimation algorithms have been extensively studied for many years. Some approaches use Random Sample Consensus (RANSAC)~\cite{Isa19} or Hough Transformation~\cite{Tian10} achieving solid results and near real-time running speed. However, such techniques consume large computational resources when dealing with dense inputs (point cloud or depth image), and require additional sensors for depth estimation. Their run-time and segmentation precision are also highly correlated with the complexity of the scene. 
Nevertheless, geometry model based plane estimation methods are indispensable for generating the ground truth, which is needed for training machine learning approaches.

Instance segmentation is the task of detecting and delineating each distinct object of interest appearing in an image. The problem of piece-wise planar region estimation from a single RGB image can be reverted to a binary-class (planar and non-planar) instance segmentation problem. 
In recent years, great progress has been made in instance segmentation with the help of convolutional neural networks (CNN). State-of-the-art two-stage approaches like FCIS~\cite{Li16} and Mask R-CNN~\cite{He17} depend on feature localization to produce masks of the objects. 
PlaneRCNN~\cite{Liu18-2} is a multi-task plane estimation model, whose detection network is built upon Mask R-CNN. It inherits not only the high accuracy advantage of the latter, but also its run-time limitations.

In this paper, we present a novel single-stage instance segmentation architecture for piece-wise planar regions estimation which reaches significantly higher frame-rates and improved segmentation accuracy against two-stage methods. In detail our contributions in this paper are:
\begin{itemize}
\item We propose PlaneSegNet, the first real-time single-stage detector for piece-wise plane segmentation. PlaneSegNet is shown in our experiments to outperform the state-of-the-art both in segmentation accuracy and run-time.
\item We improve the localization and segmentation accuracy of the network by enhancing spatial context features and introducing Fast Feature Non-maximum Suppression (FF-NMS).
\item We annotate with piece-wise plane masks and make available 70,000 images from the dataset Stanford 2D-3D-Semantics~\cite{Maarten17} using an NDT-RANSAC method.
\end{itemize}

The rest of the paper is organized as follows: First we discuss related work in Section \ref{sec:related}. Our proposed network architecture, specific improvements and ground truth generation method are discussed in Section \ref{sec:method}. The evaluation of the approach (quantitative and qualitative comparison, testing with Planar SLAM) is given in Section \ref{sec:result}. Finally, we give concluding remarks in Section \ref{sec:conclusion}.


\section{Related Work}\label{sec:related}

In this section, we summarize the relevant research done on deep learning method for piece-wise planar reconstruction and SLAM system using semantic planar regions as cues.

\subsection{Piece-wise Plane Segmentation}
PlaneNet~\cite{Liu18-1} is the first end-to-end neural network for piece-wise planar reconstruction from a single RGB image. It is built upon Dilated Residual Networks (DRNs)~\cite{Yu17} and has three prediction branches: plane parameters estimation, non-planar depth map estimation and plane segmentation. The segmentation branch starts with a pyramid pooling module followed by CRFasRNN \cite{crfasrnn_iccv2015} layers to refine the prediction. 

In PlaneRecover~\cite{Yang18}, Yang and Zhou discuss the difficulty of obtaining ground truth plane annotations in real dataset. Instead, they utilize a synthetic dataset~\cite{ros2016synthia} of urban scenes and introduce a novel plane structure-induced loss to train the network without direct supervision. However, both PlaneNet and PlaneRecover require the maximum number of planes in an image as prior. Instead of using a fully CNN network, Z. Yu et al.~\cite{yu2019single} utilize an encoder-decoder architecture to first perform semantic segmentation, and then use associative embedding model with mean shift clustering method to further segment the planar region into piece-wise plane instance. 

PlaneRCNN~\cite{Liu18-2} proposes a more effective plane segmentation branch built upon Mask R-CNN~\cite{He17}. The network shows high generalization ability across both indoor and outdoor scenes, but fails to reach real-time frame rate. Mask R-CNN based methods generate candidate region-of-interests (ROIs) and segment them sequentially. Therefore, the run-time is strongly influenced by scene complexity and object size, which further harms the frame rate robustness in real-time applications.

\subsection{Planar SLAM}
Most of the existing methods of SLAM systems are based only on points to describe the scenes and estimate the camera poses, which encounter various problems in practical application such as low-texture environments and changing light. Semantic cues are added in SLAM as geometric regularization regarding to different landmarks and optimize the geometric structure jointly, such as the distance between plane and associated 3D points, or the perpendicular plane layout of an indoor scene (walls and floor) under Manhattan assumption \cite{concha2014manhattan}. An early work~\cite{taguchi2013point} presented a RGB-D SLAM system for hand-held 3D sensor using both point and plane as primitives. Followed by the works \cite{salas2014dense, kaess2015simultaneous, hsiao2017keyframe, hosseinzadeh2018structure, zhang2019point} which tackle the problem similarly by extracting plane from depth image and optimize the poses of keyframes and landmarks (point and plane) in Bundle Adjustment (BA). More recently, \cite{yang2019monocular} employs high-level object and plane landmarks with Monocular ORB-SLAM2~\cite{mur2017orb}, the built map is dense, compact and semantically meaningful compared to the classic feature-based SLAM. Similarly, SlamCraft~\cite{Rambach19} presented an efficient planar monocular SLAM which fuses the detected plane from PlaneNet iteratively with the point cloud, resulting in higher accuracy of camera localization.

\section{METHODOLOGY}\label{sec:method}
In this section, we describe our proposed network architecture, the motivation behind our design choices, and the steps we take to reduce inference run-time.
\subsection{PlaneSegNet Overview}\label{sec:overview}

\begin{figure*}[htb]
\centering  
\includegraphics[width=0.95\textwidth]{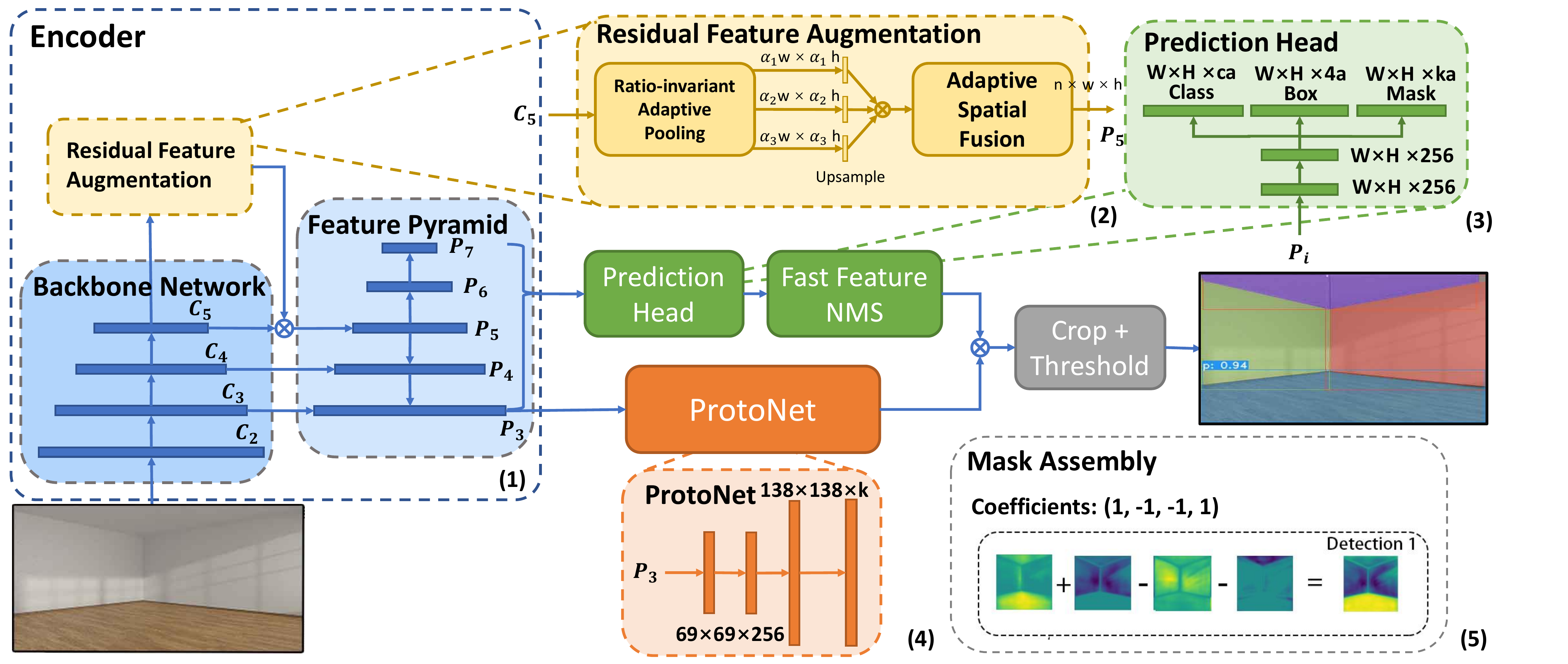}
\caption{\textbf{PlaneSegNet Architecture:} \textbf{(1) The Encoder:} a ResNet backbone combined with a Feature Pyramid Network. \textbf{(2) Residual Feature Augmentation:} an extra path from $C_{5}$ to $P_{5}$ to enhance spatial context. \textbf{(3) Prediction Heads:} provides classes, boxes and mask coefficients' prediction. \textbf{Fast Feature NMS} suppresses redundant proposals. \textbf{(4) ProtoNet:} predicts $k$ instance independent prototype masks. \textbf{(5) Assembly:} the instance masks come from a linear combination of the prototype masks and the mask coefficients.
}
\label{fig:psn}
\end{figure*}

Our proposed PlaneSegNet is build upon the YOLACT++~\cite{Bolya2019YOLACTBR} instance segmentation network with several modules optimized for the task at hand (see Figure \ref{fig:psn} for an illustration). A ResNet~\cite{He15} backbone with Feature Pyramid Network (FPN) serves as the encoder and provides a set of multi-scale feature maps $\{P_{3},P_{4},P_{5},P_{6},P_{7}\}$. Layer $C_{5}$ of the backbone is connected to the prediction layer $P_{5}$ of the FPN through a regular lateral path and an extra Residual Feature Augmentation path (see Sec. \ref{sec:augfpn}) to enhance spatial context information. The Protonet is a set of fully convolutional layers, which predicts $k$ channels for instance independent prototype masks from the feature maps $P_{3}$. The prediction heads provide $4$ bounding box regressors, $c$ class confidences and $k$ mask coefficients for $k$ prototype masks from every level of the pyramid feature maps $\{P_{3},P_{4},P_{5},P_{6},P_{7}\}$. 
We propose the Fast Feature NMS (see Sec. \ref{sec:softnms}) instead of the Fast NMS to improve the detection robustness on overlapping instances. Finally we assemble those prototype masks with coefficients prediction with a linear combination of both followed by a sigmoid non-linearity function:
\begin{equation}
M  = \sigma (P \bm{C}^{T})
\end{equation}
where $P$ is an $ h \times w \times k$ tensor of prototype masks and $\bm{C}$ is a $n \times k$ matrix of mask coefficients for $n$ instances surviving from Fast Feature NMS (see Sec. \ref{sec:softnms}) and score thresholding.

The loss function is defined as a weighted sum of the localization loss (bounding box loss), the confidence loss and the mask loss as
\begin{eqnarray}
\mathcal{L}(x,C,B,M) = \frac{1}{N}(\mathcal{L}_{conf}(x,C) + \alpha \mathcal{L}_{loc}(x,B,B_{gt}) \nonumber \\ + \beta \mathcal{L}_{mask}(x,M,M_{gt}))
\end{eqnarray}
where $N$ is the number of positive matches, and $C, B, M$ represent confidence, bounding box and mask, respectively. $\mathcal{L}_{confs}$ and $\mathcal{L}_{loc}$ are the same as the loss functions of a single-shot detector~\cite{LiuSSD15}. $\mathcal{L}_{mask}$ is the pixel-wise binary cross entropy between predicted masks and the ground truth masks.

\subsection{Piece-wise Planes as Instance Segmentation}
When we consider piece-wise planar regions as instances in scenes, our segmentation target shows particularities different from common instance segmentation tasks. In Table ~\ref{tab:plane_analyse}, we analysed the bounding box overlapping frequency (the percentage of frames with overlapping objects through the data sample), bounding box IoU distribution and the instance size distribution of COCO~\cite{lin2014microsoft} (a common multi-class instance segmentation) with the piece-wise plane annotations of ScanNet (given by~\cite{Liu18-2}) and Stanford 2D-3D-S (labeled by us).
We observed the fact that the bounding boxes of plane instances overlap in most of the frames in indoor datasets, which is much more common than that of object instances from COCO (only 0.015\%). We also found out that instances with a mask area greater than 10\% of the frame size appear more frequently in the plane annotations of ScanNet~\cite{dai2017scannet} and 2D-3D-S~\cite{Maarten17}. It is also intuitively obvious that planes in indoor scenes appear as background room structures and surfaces of foreground objects, which leads to dense and nested spatial distribution of their bounding boxes. Therefore, we are motivated to improve our approach from two aspects: 
\begin{itemize}
\item Optimize the mask quality of large size instances by compensation for spatial information loss on prediction layers. As accurate segmentation of dominant planar regions benefits our primary targeted application, visual SLAM.
\item Decrease the false suppression on overlapping bounding boxes of different instances by introducing multi-steps NMS method (Fast Feature NMS). This target shares similarity with dense pedestrian segmentation.
\end{itemize}

\begin{table}[t]
    \begin{center}
    \footnotesize
    \scalebox{0.94}{
    \begin{tabular}{|l|c|c|c|c|}
     \hline
     Dataset & Overlap(\%) & IoU(25-50\%) & IoU($\geq$50\%) &  Large-Obj.(\%)\\ \hline \hline
     COCO     & 0.015\%      & 9.01\%        & 4.55\%         &  12.27\%\\
     ScanNet  & 69.01\%      & 10.72\%       & 2.94\%         &  23.41\%\\
     2D-3D-S  & 80.08\%      & 18.11\%       & 7.18\%         &  32.84\%\\ \hline
\end{tabular}}
\end{center}
\caption{\textbf{Instance Feature Comparison} for COCO (multi-class annotations), Stanford 2D-3D-S (plane annotations) and ScanNet (plane annotations), using 2,000 random samples of each. Metrics: Overlap(\%) - percentage of frames with overlapping bounding boxes in the sample dataset, IoU in range of $(a\%, b\%)$ - percentage of overlaps with a IoU within $(a\%, b\%)$ among all overlapping bounding boxes in the sample, Large-Obj.(\%) - percentage of large instance (whose area is greater than 10\% of the frame size) among all instances in the sample.} 
\label{tab:plane_analyse}
\end{table}

\subsection{Residual Feature Augmentation}
\label{sec:augfpn}
To improve the mask quality of large size instances, we combine the low-resolution, semantically strong features with the high-resolution, semantically weak features from different depths of the backbone network via a top-down pathway and lateral connections within the Feature Pyramid Network (FPN). This feature pyramid enriches the semantics at all levels without sacrificing representational speed or memory.

Nevertheless the feature channels are reduced to 256-D by applying a convolution when building lateral connections, which causes information loss at the highest pyramid level. Furthermore, as the layer of the top down path way, $P_{5}$ only contains single scale context information. We utilize the Residual Feature Augmentation based on~\cite{Guo19} to build an extra connection from highest feature layer to its correspondent prediction layer, so that the lost information can be compensated by incorporating the spatial context information into $P_{5}$. The structural details of the Residual Feature Augmentation are illustrated in Figure \ref{fig:psn} (2).

We first perform ratio-invariant adaptive pooling on $C_{5}$ to obtain multiple context features with different scales of $(\alpha_{1} \times S, \alpha_{2} \times S,..,\alpha_{n} \times S)$, which in our case $(\alpha_{1}, \alpha_{2},\alpha_{3})=(0.1,0.2,0.3)$. Each context feature will then independently pass through a convolution layer to reduce the channel dimension to 256. After that, we upsample these features to the same scale as $C_{5}$ through bilinear interpolation and fuse them subsequently. 

The subsequent Adaptive Spatial Fusion (ASF) module combines these upsampled context features and produces spatial weight maps for each of them. Different to the original design \cite{Guo19}, we decrease the channel of spatial weight branch to its half to accelerate the network. The weights are used to generate a set of residual feature maps, which contains multi-scale context information. Subsequently, we combine them with the feature map from top-down lateral layers by summation, and perform a $3\times3$ convolution to build the beginning ($P_{5}$) of prediction layers $\{P_{3},P_{4},P_{5},P_{6},P_{7}\}$. 

\subsection{Fast Feature Non-maximum Suppression}
\label{sec:softnms}

For piece-wise plane instance in indoor scenario, bounding boxes with high overlap ($IoU\geq50\%$) do not necessarily belong to the same object in our scenario. Therefore, a classic NMS method is not optimal to solve the redundant proposals from the prediction head.

Addressing this issue, we introduce a novel method, Fast Feature Non-maximum Suppression (see Algorithm \ref{alg:nms}), which is inspired by ~\cite{salscheider2020featurenms}. Similar as Fast NMS, a $c\times n\times n$ pairwise IoU matrix $X$ is computed for the top $n$ detections sorted descending by score for each of $c$ classes (in our case, $c=1,~n=200$). The column-wise maximum $K$ is computed from the upper triangle matrix $X^{triu}$.

\begin{algorithm}[hbt]
\SetAlgoLined
\SetKwInOut{Input}{input}
\Input{ $P \leftarrow Sort(Proposals)$ with Scores, $D \leftarrow \varnothing$\;}
$ X^{triu} \leftarrow GetPairwiseIoU(P)$\;
$ K \leftarrow \max(X^{triu})$ column-wise\;
\eIf{$K_{i} \leq N_{1}$}
{$PUSH(p_{i},D)$\;}
{\uIf{$K_{i} \leq N_{2}$}
{$C^{triu} \leftarrow GetCosineSim(p, D)$\;
$S \leftarrow \max(C^{triu})$ column-wise\;
\uIf{$S_{i} \leq T$}
{$PUSH(p_{i},D)$}}
}
\Return $D$
\caption{Fast Feature NMS} \label{alg:nms}
\end{algorithm} 

If the maximal IoU value is less or equal than threshold $N_{1}$, the detection is considered to belong to different objects, and when the IoU is larger than threshold $N_{2}$ the detection is considered to belong to the same object. In our network the final mask segmentation is assembled from the prototype masks with mask coefficients prediction. Therefore, we compute the cosine similarity matrix $S$ of mask coefficients vectors between the detections $p$ whose $ IoU \in (N_{1},N_{2})$ with marked detections $d$. We set a similarity threshold $T_{1}$ to indicate which highly overlapping detection to keep for each class. The algorithm is very efficient since it only involves matrix operation. Compared to Fast NMS, our method introduce an overhead less than 1 ms.

\begin{table*}[h]
    \small
	\begin{center}
			\begin{tabular}{|l||c|c|c|p{0.55cm}<{\centering}p{0.55cm}<{\centering}p{0.55cm}<{\centering}p{0.55cm}<{\centering}p{0.55cm}<{\centering}|p{0.55cm}<{\centering}p{0.55cm}<{\centering}p{0.55cm}<{\centering}p{0.55cm}<{\centering}p{0.55cm}<{\centering}|}
				\hline
				& \multirow{2}{*}{\textbf{Backbone}} &\multirow{2}{*}{\textbf{Training}}  
				&\multirow{2}{*}{\textbf{FPS}} 
				&\multicolumn{5}{c|}{\textbf{TUM RGB-D} \cite{sturm12iros}} 
				& \multicolumn{5}{c|}{\textbf{NYU-V2} \cite{silberman2012indoor}}\\
				& & & & $AP^{b}_{50}$ &$AP^{m}_{50}$  & $VOI\downarrow$ & $RI$ & $SC$ & $AP^{b}_{50}$ & $AP^{m}_{50}$ & $VOI\downarrow$ & $RI$ & $SC$ \\ 
				\hline
				\hline
				\textbf{PlaneRCNN} \cite{Liu18-2}& ResNet101 & ScanNet & 3.0
				& 37.13  &  31.93 &  \textbf{1.502}  &  0.746  &  \textbf{0.652}
				& 20.02  &  18.89 &  2.861  &  0.724  &  0.458\\
				\hline
				\textbf{PlaneSegNet-101} & ResNet101 & ScanNet &  \textbf{30.2}
				&  34.82  &  33.80  &  1.759  &  0.733  &  0.602
				&  26.36  &  21.69  &  2.794  &  0.768  &  0.481  \\
				\hline
				\textbf{PlaneSegNet-101} & ResNet101 & 2D-3D-S &  \textbf{30.2}
				&  \textbf{43.97}  &  \textbf{40.52}  &  1.645  &  \textbf{0.748}  &  0.628
				&  \textbf{32.78}  &  \textbf{21.74}  &  \textbf{2.524}  &  \textbf{0.774}  &  \textbf{0.521} \\
				\hline \hline
				\textbf{Baseline (Yolact++)} & ResNet101 & 2D-3D-S &  31.5
				&  41.62  &  37.64  &  1.724  &  0.734  &  0.607 
				&  32.02  &  20.83  &  2.591  &  0.766  &  0.506   \\
				\hline
				\textbf{Baseline + FF-NMS} & ResNet101 & 2D-3D-S & 31.4
				&  41.80  &  37.81  &  1.708  &  0.737  &  0.612 
				&  32.39  &  20.86  &  2.573  &  0.769  &  0.511   \\\hline
				\textbf{Baseline + Aug-FPN} & ResNet101 & 2D-3D-S & 30.7   
				&  43.72  &  40.33  &  1.673  &  0.744  &  0.620
				&  32.14  &  21.55  &  2.557  &  0.770  &  0.513\\\hline
			\end{tabular}
	\end{center}
	\caption{\textbf{Quantitative comparison of our approach against other state-of-the-art methods and ablation study of our contributions} on dataset TUM RGB-D~\cite{sturm12iros} (about 1,300 samples), dataset NYU V2~\cite{silberman2012indoor} evaluated in terms of mask Average Precision ($AP^{m}$), bounding box Average Precision ($AP^{b}$), Variation of Information (VOI), Rand Index (RI), Segmentation Covering (SC) and run-time (FPS).} 
	\label{tab:results}
\end{table*}


\section{Experiments and Results}
\label{sec:result}
There are three different experiments conducted in our work: first we compare our PlaneSegNet with other state-of-the-art quantitatively and qualitatively. Second we ablate our added components in terms of their performance and the trade-off between run-time and accuracy. Third we present an application of our network combined with a monocular planar SLAM system. 

\subsection{Benchmarking and Experimental Setup} \label{sec:gt}

\begin{figure}[t]
  	\centering  
	\includegraphics[width=0.46\textwidth]{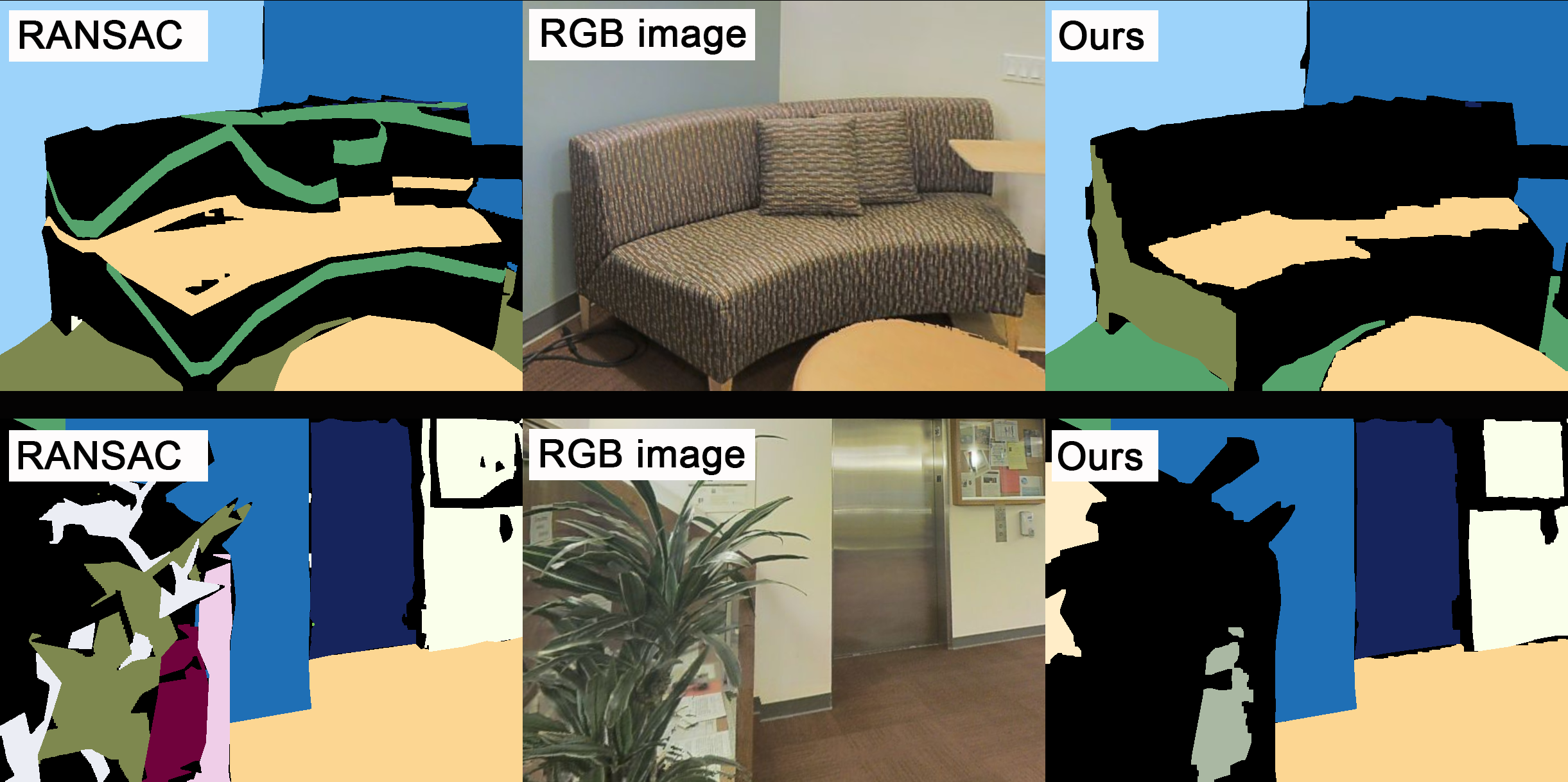}
    \caption{\textbf{Comparison of ground truth generation of the applied NDT-RANSAC against simple RANSAC:} in the first column, the example (top row) shows the simple RANSAC failed due to the absence of normal direction of points, or (bottom row) failed by attempting to fit planes in curve surface. Raw RGB images are extracted from 2D-3D-S~\cite{Maarten17}.}
   {\label{fig:falied_ransac}}  
\end{figure}

We present a new benchmark from RGB-D image on dataset 2D-3D-S~\cite{Maarten17} with improved RANSAC method based on Normal Distribution Transformation (NDT) ~\cite{Magnusson20073DNDT}, following the work of Li et al.~\cite{Li17}. Comparing to simple RANSAC, the approach shows better performance as illustrated in Figure \ref{fig:falied_ransac}. The main reasons are: 
the method considers both the normal direction and the location of NDT cells while estimating planes, in order to minimize incorrect fitting on step-wise object, and non-planar cells are filtered out with a threshold before RANSAC. Thus, we avoid hard-fitting on curve surface.

\begin{figure*}[h]
\centering
\includegraphics[width=0.97\textwidth]{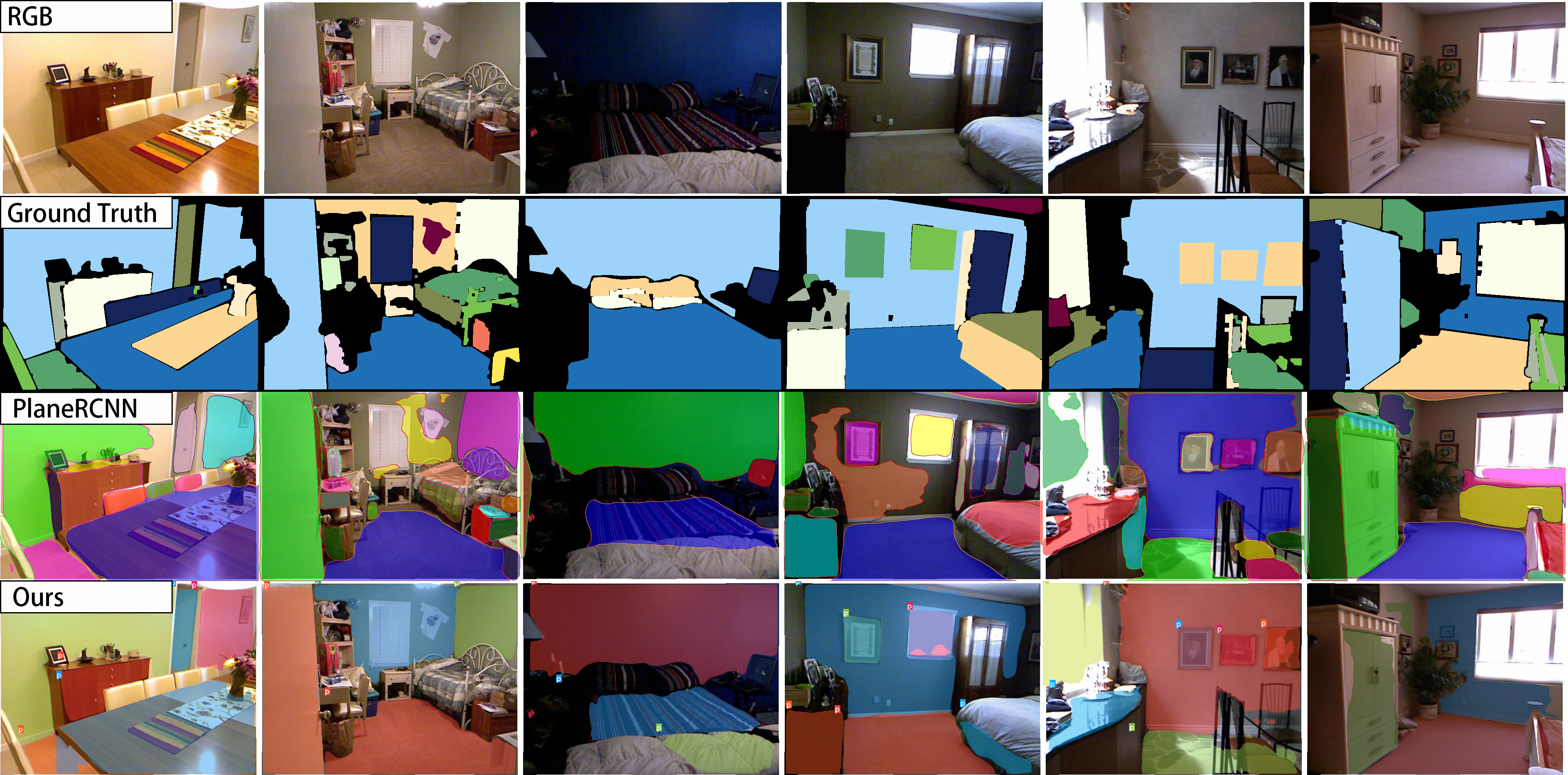}
\caption{\textbf{Qualitative results of instance planar segmentation of challenging cases} on dataset NYU v2~\cite{silberman2012indoor}. Notice that NYU dataset does not have planar semantic ground truth, the ground truth illustrated here was generated by our method (Sec.~\ref{sec:gt}). The result shows our method provides in general more compact and more precise semantic masks, especially in favor of the plane boundary.}
\label{fig:results}
\end{figure*}

We implemented our PlaneSegNet using PyTorch, and use ResNet-101~\cite{He15} as backbone with ImageNet~\cite{Deng09} pretrained weights~\cite{He15}. Newly added layers are initialized as random filters sampled from a normal distribution with zero mean and variance. We trained our network end-to-end on a NVIDIA RTX 3090 for 15 epochs with Adam~\cite{kingma2014adam} optimizer on the dataset 2D-3D-S~\cite{Maarten17} with piece-wise plane annotations, which consists of about 70,000 images. 
For comparison purposes we also trained a model of our network on ScanNet~\cite{dai2017scannet} (70,000 image samples, with the annotation provided by Liu et al.~\cite{Liu18-2}). 

Considering a major application of our work is visual-SLAM, we additionally utilize motion blur and Gaussian noise as data augmentation methods, to enhance our network's robustness against noisy input images.

\subsection{Instance Planar Segmentation Results}
\label{sec:eval}

In Table \ref{tab:results}, we compare our approach to two piece-wise plane reconstruction networks: PlaneNet \cite{Liu18-1} and PlaneRCNN \cite{Liu18-2}. We test on two dataset previously not used for this purpose, namely TUM RGB-D \cite{sturm12iros} (average 2.92 planes per image) and the NYU-V2 labeled subset~\cite{silberman2012indoor} (average 11.63 planes per image) with the plane segmentation ground truth generated by us. In order to exclude the influence of different training datasets, we also give the result of our PlaneSegNet trained on ScanNet with the annotation given by Liu et al. \cite{Liu18-2}.  
We evaluate the run-time in terms of frame per second (FPS), the detection using bounding box average precision ($AP^{b}$) and the segmentation performance using mask average precision ($AP^{m}$), VOI, RI and SC. The frame rate of all methods is tested with an indoor scene video with image size of $640\times480$, running all models on the same GPU (GTX 1080 Ti). Since PlaneNet and PlaneRCNN are multi-task networks, we disable their depth estimation and plane parameter estimation branches during the run-time testing. Our PlaneSegNet-101 trained on 2D-3D-S has the best result in terms of mask and bounding box Average Precision, while PlaneRCNN shows the best results in terms of VOI, RI and SC, however these metrics are strictly designed for evaluating semantic segmentation, instead of instance segmentation.

The qualitative results of planar segmentation are illustrated in Figure \ref{fig:results}, where PlaneRCNN shows higher detecting ability on small objects, but suffers from the tendency to generalize as many positive detections as possible, which sometimes leads to false positive results. It also results in the average precision of PlaneRCNN being much lower than our method on the TUM RGB-D dataset, because PlaneRCNN divides many large scale planes into individual small instances which will not be considered as matched prediction when computing AP. Meanwhile our method provides better mask boundary quality on dominant planes in the image and maintains the instance completeness. 

\begin{figure}[t]
\centering
\subfloat[Mask quality.]{
\includegraphics[width=0.85\linewidth]{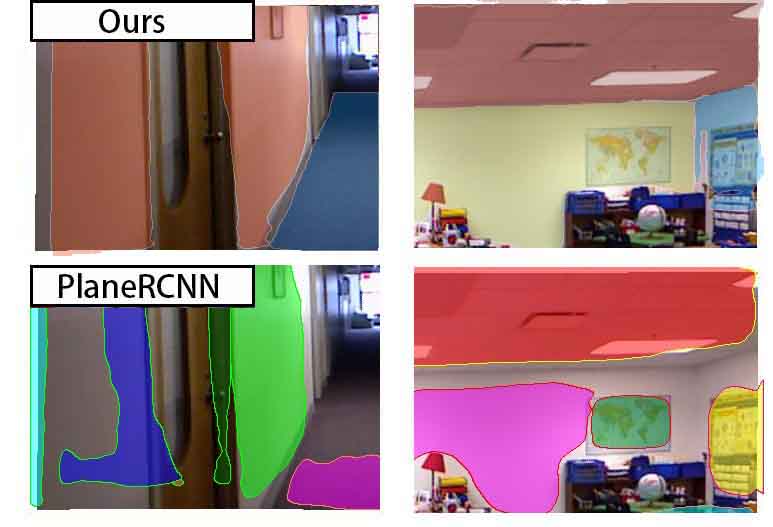}}\\
\subfloat[Failure cases.]{
\includegraphics[width=0.85\linewidth]{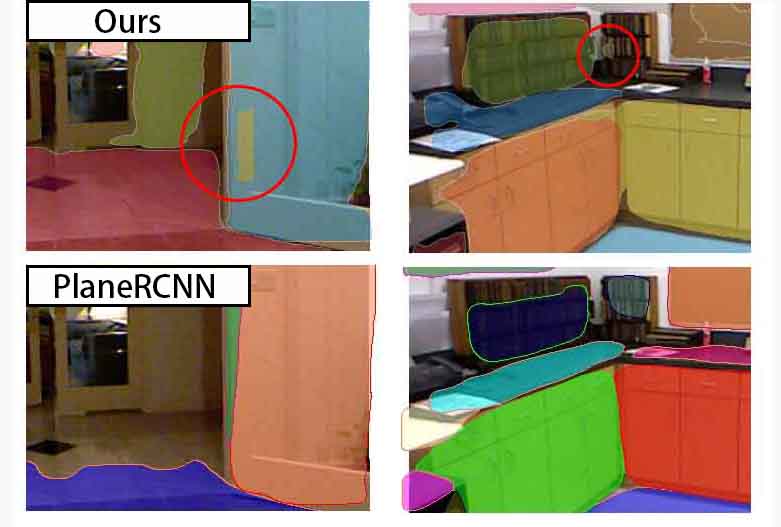}}
\caption{\textbf{Segmentation quality and failure cases.} In (a) we illustrate the quality. Our segmentation quality is better than PlaneRCNN in terms of the mask boundary completeness. In (b) we show some failure cases of feature leakage.}
\label{fig:results_details}
\end{figure}

For visual SLAM applications, high quality segmentation prediction for large-size dominant planes significantly benefits the tracking accuracy. PlaneRCNN is based on two-stage segmentation method and local mask segmentation inside ROIs, while our PlaneSegNet is based on global prototype mask assembly. This improves our method's prediction accuracy on large scale instances (dominant plane), as shown in Figure~\ref{fig:results_details} (a). Some failure cases of PlaneSegNet can also be seen in Figure~\ref{fig:results_details} (b). The reason of these failures is that the prediction masks are cropped after assembly and no further suppression module is involved to filter out the noise from the cropped region. 

\subsection{Ablation Study}
To better understand how the components of our model contribute to the overall performance, in Table~\ref{tab:results} we perform an ablation study by changing various components of our model. As shown in the Table, the Residual Feature Augmentation module enhances the spatial context and improves the segmentation and bounding box prediction accuracy in terms of all the metrics. By introducing the Fast Feature NMS the prediction accuracy of bounding box estimation is improved more obvious than mask segmentation. We also trained our network with the same amount of data and iteration on 2D-3D-S and ScanNet, to compare and to prove that our ground truth generation method is more reliable and precise than the method used in PlaneRCNN. 

\subsection{Incorporating with Planar SLAM}
\label{sec:application}

\begin{figure}[t]
\centering
\includegraphics[width=0.5\textwidth]{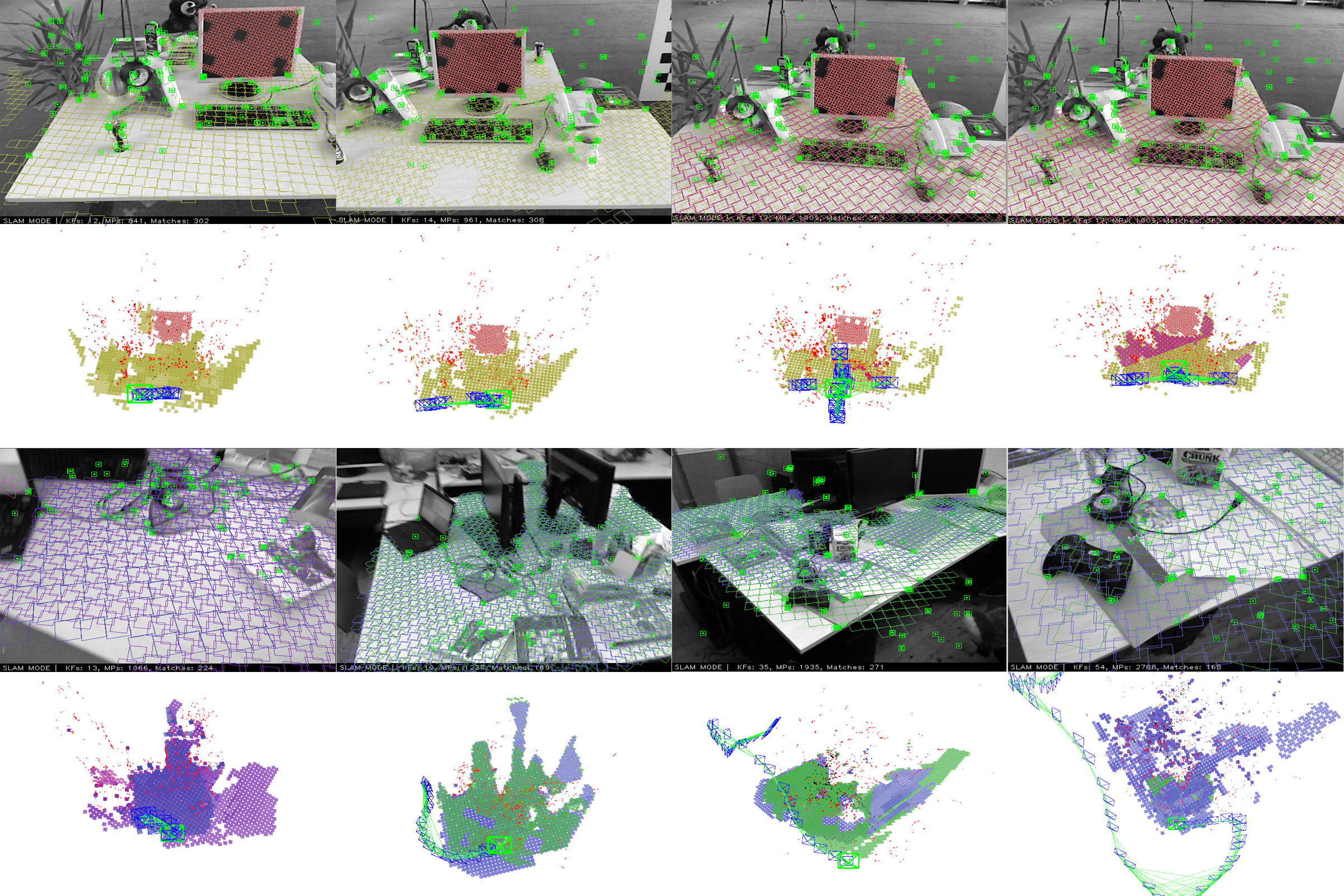}
\caption{\textbf{Qualitative results of dense planar map} generated from SlamCraft~\cite{Rambach19} using our PlaneSegNet. The two image sequences are fr2\_xyz and fr1\_desk from TUM RGB-D~\cite{sturm12iros}.}
\label{fig:slam}
\end{figure}

In this work, we employ SlamCraft~\cite{Rambach19} which is an efficient planar SLAM using monocular image sequence and generating a compact surfel map representation based on semantic cues. We replace the plane segmentation network used in SlamCraft (PlaneNet) with our PlaneSegNet which is about 5 times faster. We present the comparison of tracking accuracy results from  Monocular ORB-SLAM2, SlamCraft+Ours and SlamCraft+PlaneNet in Table \ref{tab:slamtrack}, respectively, as well as the qualitative results of dense planar map in Figure~\ref{fig:slam}. Our work combined with SlamCraft shows better results in terms of providing robust segmentation mask and improved localization accuracy of the camera. 

\begin{table}[t]
	\begin{center}
	\footnotesize
			\begin{tabular}{|c|c|c|c|c|c|c|}
				\hline
				\multicolumn{7}{|c|}{Absolute KeyFrame Trajectory RMSE (cm)}\\
				\hline
				\textbf{Seq.}&\multicolumn{2}{|c|}{\textbf{ORB-SLAM2}}&\multicolumn{2}{|c|}{\textbf{SlamCraft}}&\multicolumn{2}{|c|}{\textbf{SlamCraft+Ours}}\\
				\hline
				 \# & Mean & Med.& Mean & Med.& Mean & Med.\\
				\hline
				\hline
				fr1 xyz & 0.76 & 0.62 & \textbf{0.66} & \textbf{0.61}& 0.72 & 0.62 \\
				\hline
			    fr2 xyz & 0.13 & 0.14 & 0.12 & 0.12& \textbf{0.11} & \textbf{0.12} \\
				\hline
				fr1 desk & 0.68 & 0.68 & 0.60 & 0.47& \textbf{0.58} & \textbf{0.46} \\
				\hline
				fr2 desk & 0.88 & 0.90 & 0.89 & 0.91& \textbf{0.75} & \textbf{0.72} \\  
				\hline
			     fr2 sit xyz & 0.32 & 0.32 & 0.31 & 0.31& \textbf{0.31} & \textbf{0.30} \\
				\hline
			    fr3 str tex far  & 0.86 & 0.87 & 0.84 & 0.84& \textbf{0.76} & \textbf{0.68} \\
				\hline
\end{tabular}
	\end{center}
	\caption{\textbf{Comparison of Monocular ORB-SLAM2~\cite{mur2017orb} to SlamCraft~\cite{Rambach19}} in terms of trajectory RMSE (Root Mean Square Error) on dataset TUM RGB-D~\cite{sturm12iros}. The original SlamCraft employs PlaneNet as their planar detector, while we replace it with our PlaneSegNet (SlamCraft+Ours) which improves accuracy in 5 out of the 6 image sequences.}
	\label{tab:slamtrack}
\end{table}

\section{Conclusion}
\label{sec:conclusion}
We present the first real-time single-stage instance segmentation method for piece-wise plane estimation. By optimizing the network structure and hyper parameters, we achieve a balance between accuracy and frame rate. Experiments against PlaneRCNN  and PlaneNet demonstrated the effectiveness of our approach. Our method shows better segmentation quality in large scale planar regions while being real-time capable. Additionally, an application of our method on planar visual SLAM is presented in this work. Our approach still has some limitations in complex scenes with multiple scale instances. Further improving the bounding box localization accuracy and exploring more effective mask assembly solutions with acceptable run-time overhead will be interesting directions for future work.

\clearpage
\bibliographystyle{ieee}
\bibliography{egbib}

\end{document}